\documentclass[runningheads]{llncs}

\usepackage{eccv}

\usepackage{eccvabbrv}

\usepackage{graphicx}
\usepackage{booktabs}

\usepackage[accsupp]{axessibility}  

\usepackage{hyperref}

\usepackage{wrapfig}
\usepackage{subcaption}
\usepackage[table]{xcolor}
\usepackage{tabularx}
\usepackage{multirow}

\usepackage{orcidlink}
\usepackage{xcolor}

\definecolor{TiffanyBlue}{RGB}{129,216,208}
\definecolor{AncoraRed}{RGB}{128,0,32}
\definecolor{KleinBlue}{RGB}{0,47,167}

\begin{document}

\title{Recolour What Matters: Region-Aware Colour Editing via Token-Level Diffusion} 

\titlerunning{Recolour What Matters}

\author{Yuqi Yang\inst{1, 2} \and
Dongliang Chang\inst{1, 2}\and
Yijia Ling\inst{1, 2} \\
Ruoyi Du\inst{1}\and
Zhanyu Ma\inst{1, 2}
}

\authorrunning{Y. Yang et al.}

\institute{Beijing University of Posts and Telecommunications, Beijing 100876, China \and
Beijing Key Laboratory of Multimodal Data Intelligent Perception and Governance, Beijing 100876, China\\
\email{\{yangyuqi, changdongliang, yijialing, duruoyi, mazhanyu\}@bupt.edu.cn}\\
\url{https://yangyuqi317.github.io/ColourCrafter.github.io/} }

\maketitle

\begin{figure}
    \centering
    \includegraphics[width=0.8\linewidth]{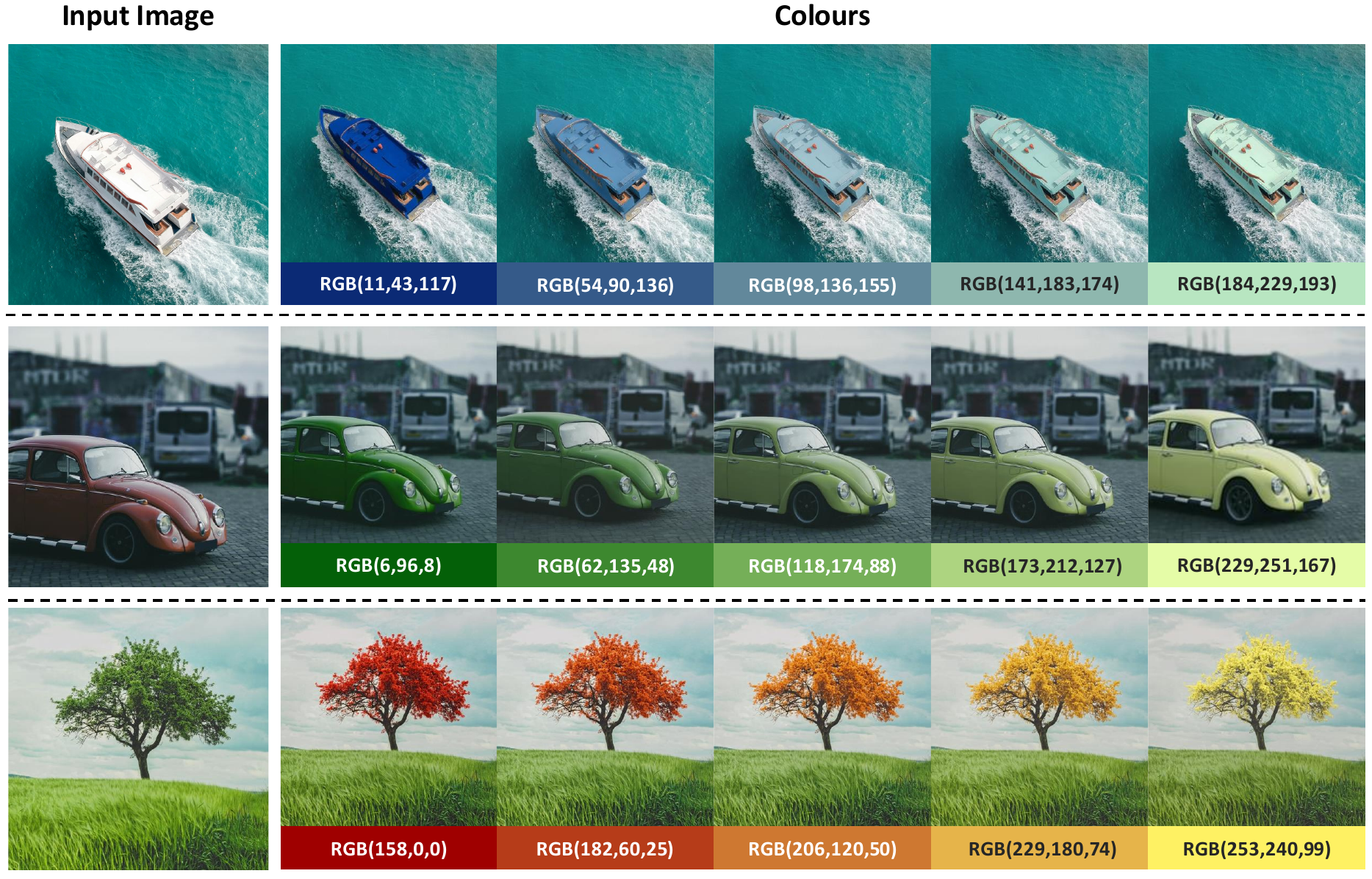}
    \captionof{figure}{{\bf Editing results of ColourCrafter under varying reference colours.} 
    Each row shows the input image and its edited outputs conditioned on different RGB references. 
    As the reference colours vary smoothly from left to right, ColourCrafter produces continuous and precise recolouring with consistent structure and texture.}
    \label{fig:fig1}
    \vspace{-25pt}
\end{figure}

\begin{abstract}
Colour is one of the most perceptually salient yet least controllable attributes in image generation.
Although recent diffusion models can modify object colours from user instructions, their results often deviate from the intended hue, especially for fine-grained and local edits.
Early text-driven methods rely on discrete language descriptions that cannot accurately represent continuous chromatic variations.
To overcome this limitation, we propose ColourCrafter, a unified diffusion framework that transforms colour editing from global tone transfer into a structured, region-aware generation process.
Unlike traditional colour-driven methods, ColourCrafter performs token-level fusion of RGB colour tokens and image tokens in latent space, selectively propagating colour information to semantically relevant regions while preserving structural fidelity.
A perceptual Lab-space Loss further enhances pixel-level precision by decoupling luminance and chrominance and constraining edits within masked areas.
Additionally, we build ColourfulSet, a large-scale dataset of high-quality image pairs with continuous and diverse colour variations.
Extensive experiments demonstrate that ColourCrafter achieves state-of-the-art colour accuracy, controllability and perceptual fidelity in fine-grained colour editing. Our project is available at \url{https://yangyuqi317.github.io/ColourCrafter.github.io/}.
  {}
\end{abstract}

\section{Introduction}
\label{sec:intro}

{Colour editing remains one of the most perceptually critical yet technically under-constrained problems in image generation~\cite{chang2023lcad, chang2022coder, chang2023coins}.}
Despite rapid progress in diffusion- and transformer-based generative models, 
{achieving fine-grained, structure-preserving, and perceptually accurate colour control remains fundamentally challenging~\cite{ruiz2023dreambooth, zhang2023sine, butt2024colorpeel, shum2025color, Dalva2025FluxSpace}.}
{Colour is not merely a stylistic attribute; it is a dominant perceptual signal that directly influences object identity, material perception, and visual realism, particularly in domains such as fashion~\cite{han2023fashionsap, huang2025text}, product design~\cite{hou2025gencolor}, and artistic creation~\cite{muratbekova2024color, shi2025fonts}. However, existing generative systems often fail to reproduce user-specified hues precisely, particularly when localised edits and chromatic consistency are required.}

Early colour editing methods are predominantly text-driven, {where target hues are described through natural language prompts}~\cite{brooks2023instructpix2pix, dong2024chromafusionnet, tsai2025color, kawar2023imagic, zhou2025fireedit, yin2025training}.
{These approaches project colour descriptions into semantic embedding spaces to guide generation.}
However, language is inherently discrete, whereas colour varies continuously in perception~\cite{butt2024colorpeel}.
Even with a rich vocabulary such as \textcolor{TiffanyBlue}{Tiffany blue}, \textcolor{AncoraRed}{Ancora red} or \textcolor{KleinBlue}{Klein blue}, it is difficult to capture subtle chromatic transitions, {and interpolation in text embedding space} often lead to discontinuous or inaccurate colour changes~\cite{tsai2025color}.
This {discreteness} fundamentally limits {smooth and controllable colour manipulation.}

{
To overcome the limitations of linguistic abstraction, recent works replace colour words with explicit colour patches or reference images~\cite{butt2024colorpeel, shum2025color, lei2025stylestudio}. 
By injecting continuous chromatic signals, these colour-driven approaches offer a more direct conditioning mechanism. 
However, most existing methods introduce colour cues as global style features within the diffusion or latent space~\cite{ye2023ip-adapter, garifullin2025materialfusion, liu2025samam, Laria2026Leveraging, lobashev2025color, rowles2024ipadapter}. 
Without establishing explicit correspondence between reference colours and target object regions, colour transfer becomes ambiguous: the generated image often inherits the overall tone of the reference rather than reproducing the precise target hue. 
Such global conditioning frequently introduces colour drift, texture distortion, or unintended structural alteration.}

{
Although several works~\cite{liang2025control, qiu2025exploring} incorporate spatial masks to constrain the extent of colour propagation, masking primarily addresses \emph{where} to recolour rather than \emph{how precisely} to reproduce the target hue. 
In the absence of explicit chromatic alignment mechanisms, even masked methods may exhibit subtle hue deviation or chromatic bias within edited regions. 
In summary, existing approaches either lack semantic localisation or fail to establish pixel-level chromatic correspondence, making precise and structure-preserving colour editing inherently difficult.}

{
Beyond modelling limitations, the scarcity of datasets containing aligned, instance-level colour variations further impedes progress. 
Without paired colour variants of the same object, models struggle to disentangle semantic identity from chromatic change, preventing reliable learning of colour–pixel correspondence.}

\begin{figure}[t]
    \centering
    \includegraphics[width=0.85\linewidth]{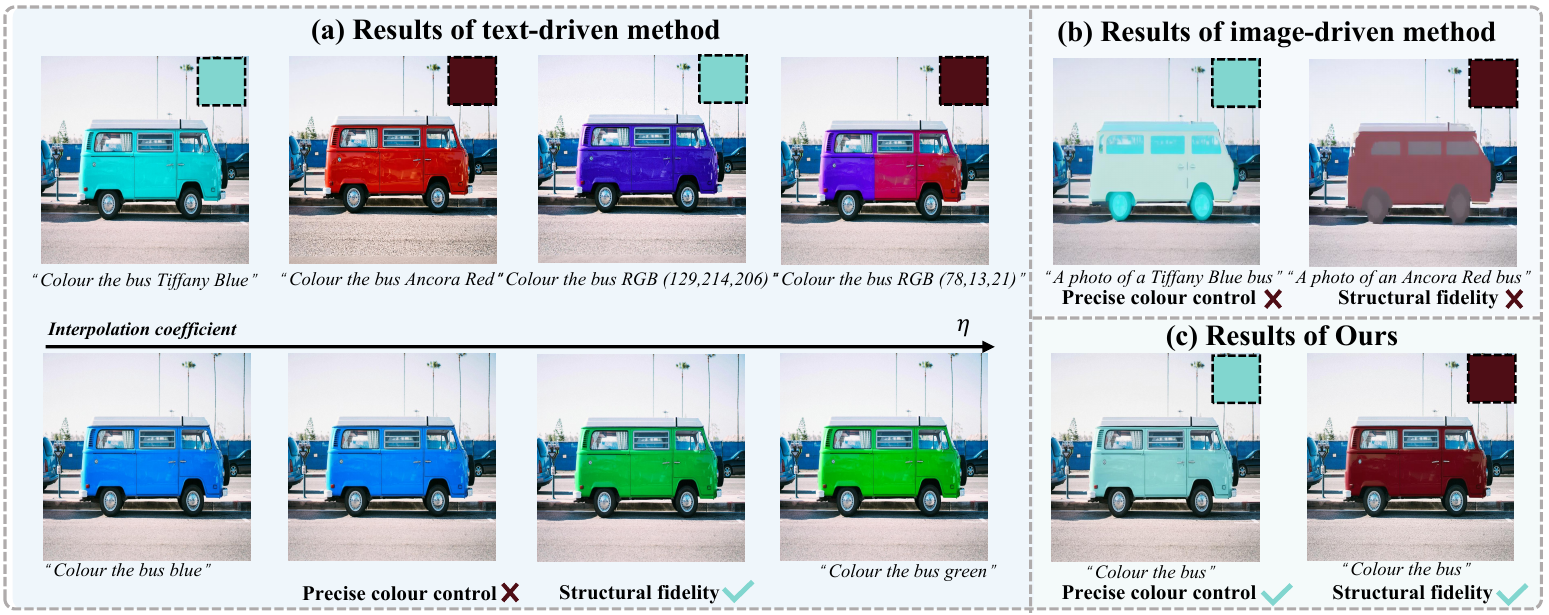}
    \vspace{-5pt}
    \caption{\textbf{Comparison of different colour editing methods.} (a) shows the results of natural language description (first row) and text embedding interpolation (second row); (b) displays the image-driven results using IP-Adapter. In comparison, our proposed method achieves both precise colour control and high structural fidelity.
}
    \label{fig:interp}
    \vspace{-20pt}
\end{figure}

To address these challenges, we propose {\bf ColourCrafter}, a unified colour-image-guided framework which achieves fine-grained and structure-preserving colour editing.
Unlike previous colour-driven approaches that apply global tone transfer, ColourCrafter performs token-level fusion of colour and image features within the diffusion process.
RGB colour patches are encoded as explicit colour tokens and fused with image tokens in latent space, enabling the model to propagate colour information selectively to relevant regions while preserving texture and structural fidelity.
This design transforms colour control from global tone transfer into a structured, region-aware generation process, allowing the model to understand \emph{what} to recolour from textual semantics and { \emph{how precisely} to recolour it from continuous RGB references. }
To further ensure perceptual fidelity, we introduce a {masked} Lab-space Loss that decouples luminance and chrominance, constraining colour deviations within edited regions while preserving unedited content.
Together, these components explicitly unify semantic localisation and continuous chromatic precision in a single diffusion framework, overcoming the global drift and structural degradation inherent to traditional colour-driven methods.

To support model training and evaluation, we construct ColourfulSet, a large-scale dataset containing high-resolution images {with diverse, instance-aligned colour variations. }
ColourfulSet significantly expands both scale and chromatic diversity and, to our knowledge, is the first dataset 
{specifically designed for continuous and locally controllable colour editing. 
By providing explicit colour–region correspondences, ColourfulSet enables supervised learning of chromatic alignment and establishes a benchmark for precise recolouring.}

In summary, our main contributions are as follows:
(1) We propose ColourCrafter, a unified diffusion framework that combines semantic localisation from text guidance with token-level RGB conditioning, transforming colour editing from global tone transfer to region-aware generation.
(2) We design a {masked} Lab-space Loss that enforces perceptual consistency and pixel-level precision by decoupling luminance and chrominance within edited regions.
(3) We build ColourfulSet, a large-scale dataset with continuous and diverse colour variations emphasising local controllability, providing a solid foundation for training and benchmarking fine-grained colour editing models.

\vspace{-10pt}
\section{Related Work}
\label{sec:related_work}

\subsection{Diffusion Models for Image Editing}
\label{subsec:diffusion_models}

Diffusion models have rapidly become the dominant paradigm for image generation~\cite{du2024demofusion, xiao2025omnigen, tan2025ominicontrol, wang2025fairhuman} and editing~\cite{brooks2023instructpix2pix, kawar2023imagic, guo2024focus}.  
Early studies relied on DDIM sampling~\cite{song2020denoising} and U-Net backbones~\cite{ronneberger2015u}, with pre-trained models such as Stable Diffusion~\cite{rombach2022high} showing strong capabilities in both synthesis and manipulation~\cite{brack2024ledits++, guo2024focus, lo2024distraction, yin2024benchmarking}.  
Prompt-to-Prompt (P2P)~\cite{hertz2022prompt} enabled controllable editing by aligning spatial layouts with textual prompts through cross-attention, while InstructPix2Pix~\cite{brooks2023instructpix2pix} extended this idea using instruction tuning for flexible text-guided manipulation.  
To move beyond explicit inversion, Imagic~\cite{kawar2023imagic} optimised text embeddings to align the original appearance with target semantics.  
More recent advances combining Flow Matching~\cite{lipman2022flow} and large-scale transformers, such as the Diffusion Transformer (DiT)~\cite{peebles2023scalable} and FLUX~\cite{flux2024}, have further enhanced image fidelity and multimodal controllability. {Meanwhile, Omni-Control~\cite{tan2025ominicontrol} explored an effective strategy for integrating image conditioning into DiT architectures, significantly enhancing conditional control in diffusion models.}

Despite these advances, most diffusion-based editors remain limited in precise colour control~\cite{park2025style, xu2025insightedit, kulikov2025flowedit, jiao2025uniedit}.  
They excel at semantic alignment but typically treat colour as a global style attribute rather than a spatially local and continuous feature.  
This motivates our work toward a unified diffusion framework that integrates semantic localisation and chromatic precision at the token level.

\vspace{-10pt}
\subsection{Colour Control}
\label{subsec:color_control}
\vspace{-3pt}

Colour editing is one of the most perceptually intuitive yet technically demanding tasks in image manipulation.  
Existing approaches are mainly text-driven~\cite{brooks2023instructpix2pix, guo2024focus, tsai2025color} or image-driven~\cite{butt2024colorpeel, shum2025color, ye2023ip-adapter, lei2025stylestudio}.  



{ColorCtrl~\cite{yin2025training} describe target colours using natural language and map textual descriptions into embedding spaces as generation conditions.
Although Tsai et al.~\cite{tsai2025color}  approximate hue variations by interpolating between text embeddings, the discrete nature of language prevents smooth and continuous colour control. 
Dong et al.~\cite{dong2024chromafusionnet} further attempted to encode RGB colour patches into prompt embeddings to enhance colour representation, yet these methods still depend on linguistic abstraction rather than explicit visual cues.

Building upon this idea, ColorPeel~\cite{butt2024colorpeel}, ColorWave~\cite{Laria2026Leveraging} and other image-driven methods~\cite{lei2025stylestudio, ye2023ip-adapter,shi2025fonts}  extend text-driven frameworks by replacing colour words with colour images, using visual references as colour conditions for generation. 
While this substitution provides continuous and measurable chromatic cues, image-conditioning approaches represented by IP-Adapter typically adopt a global conditioning injection mechanism. 
Consequently, such methods often shift overall tone, distort texture, and lack spatially localised control~\cite{Laria2026Leveraging}.
Although Control-Color~\cite{liang2025control} constrains editing regions through hint annotations, it still exhibits limitations in colour consistency and precise colour mapping.}
In essence, text-driven methods provide semantic localisation without continuity, whereas image-driven methods offer continuity without localisation. 

\textbf{ColourCrafter} bridges this gap by combining text-guided semantics and explicit RGB conditioning within a unified diffusion framework.  
Through token-level fusion of colour and image features in latent space, it achieves structured, region-aware, and perceptually consistent colour editing.

\begin{figure*}[t]
    \centering
    \includegraphics[width=0.95\linewidth]{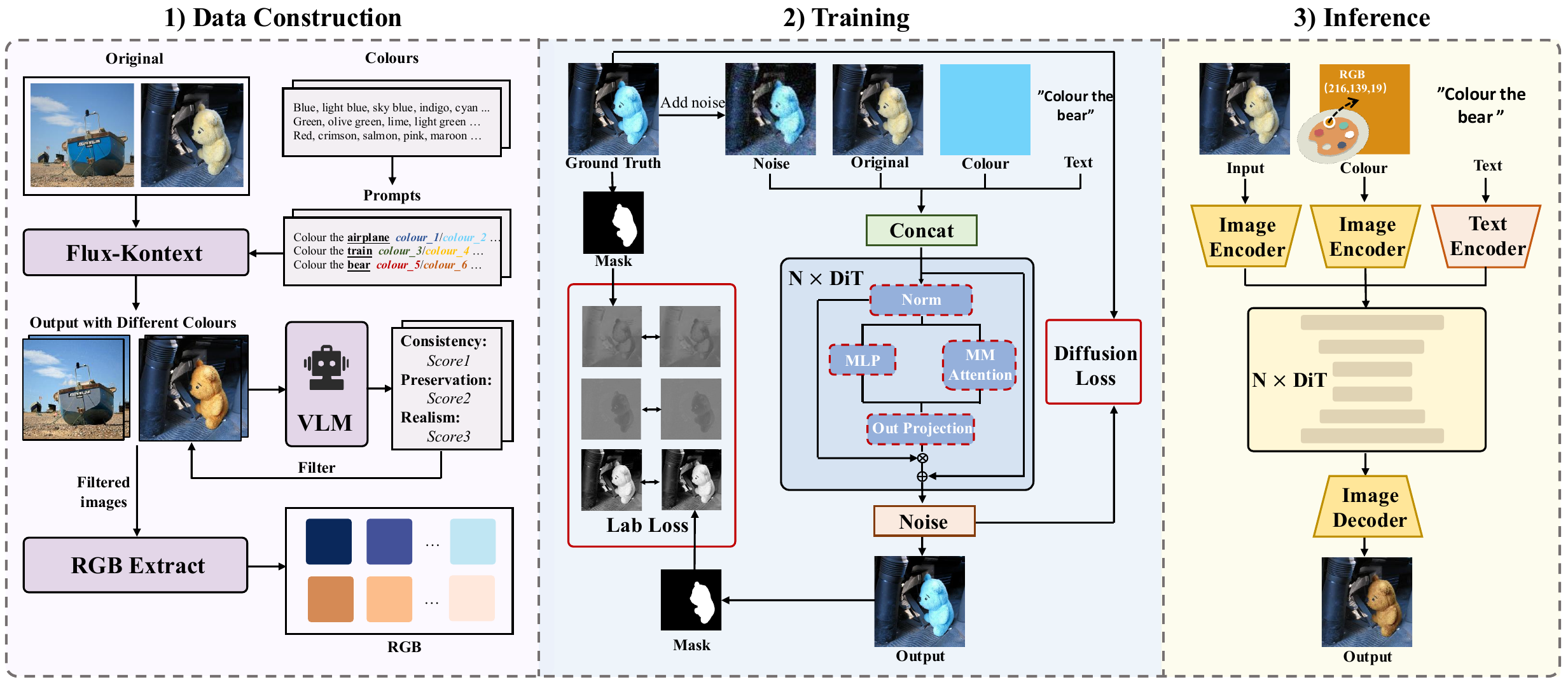}
    \caption{{\bf Overview of the ColourCrafter pipeline.} 
    (1) {\bf Dataset construction:} Using Flux.1-Kontext, we generate diverse image–colour pairs and employ a Vision–Language Model (VLM) to filter samples for consistency, fidelity, and realism. The corresponding RGB references are extracted to build the high-quality dataset \textbf{ColourfulSet}. 
    (2) {\bf Training:} The original image, target colour reference, and text prompt are jointly fed into the diffusion model, which is optimised with both Diffusion and Lab-space losses to enhance chromatic accuracy and perceptual consistency. 
    (3) {\bf Inference:} Given an input image, a RGB reference, and a prompt, ColourCrafter performs fine-grained, structure-preserving, and perceptually natural colour editing.}
    
    \label{fig:method}
    \vspace{-20pt}
\end{figure*}

\vspace{-10pt}
\section{Method}
\label{sec:method}
\vspace{-10pt}
\subsection{Preliminary}
\label{subsec:preliminary}

Our framework builds upon \textbf{Flux.1-Kontext}~\cite{labs2025flux1kontextflowmatching}, which combines Diffusion Transformer (DiT)~\cite{peebles2023scalable} with Flow Matching~\cite{lipman2022flow} for efficient and high-quality image generation.  
Flux.1-Kontext employs a transformer-based denoiser and uses Rectified Flow Matching (RFM) to stabilise training.  

Given a noisy image $x$, a reference image $I$, and a text prompt $T$, the model encodes the reference image $I$ and text $T$ into tokens $C_I$ and $C_T$, and uses the noisy image $x$ as the target token sequence.
The reference tokens are appended to the noisy image tokens $[x; C_I]$, forming a unified input sequence.  
To distinguish the spatial structures of noisy and reference images, a \textbf{3D Rotary Position Embedding (3D RoPE)}~\cite{su2024roformer} introduces an additional depth dimension, token position $u=(t,h,w)$, and $u_{C_I}=(1,h,w), u_x=(0,h,w)$,
where $(h,w)$ denotes spatial position, $u_{C_I}$represents the positional encoding of ${C_I}$, and $u_{x}$represents the positional encoding of ${x}$.
The RFM objective learns the velocity field between the noisy and clean samples:
\begin{equation}
L_{\theta}=\mathbb{E}_{t,x,C_I,C_T}\!\left[\|v_{\theta}(z_t,t,C_I,C_T)-(\epsilon-x)\|_2^2\right],
\end{equation}
where $z_t=(1-t)x+t\epsilon$, $\epsilon\!\sim\!\mathcal{N}(0,1)$.  

Although Flux.1-Kontext performs outstandingly in semantic understanding and cross-modal association, its colour editing relies on natural language descriptions, making it difficult to achieve precise control over local and fine-grained colour details. This limitation motivates our ColourCrafter design based on Flux.1-Kontext framework.

\vspace{-10pt}
\subsection{ColourCrafter}
\label{subsec:colorcrafter}

To address this limitation, we introduce \textbf{ColourCrafter}, a unified diffusion framework for fine-grained and structure-preserving colour editing. ColourCrafter bridges semantic localisation from text guidance and continuous chromatic precision from explicit RGB references, enabling controllable and perceptually consistent colour propagation.

\vspace{-10pt}
\subsubsection{Latent-space Fusion}
\label{subsubsec:latent-space}

Given a text instruction $T$, an original image $I_{orig}$, and a colour reference $I_{colour}$, the text is encoded by a frozen text encoder $\mathcal{E}_T$ into tokens $C_T$.  
Both $I_{orig}$ and $I_{colour}$ are mapped into latent representations via frozen VAEs, producing $C_{I_o}$ and $C_{I_c}$.  
These are concatenated to form a unified context:
\begin{equation}
C_I=\text{Concat}(C_{I_o},C_{I_c}).
\end{equation}

To differentiate between token types during attention computation, the noise tokens are assigned a time offset of $t\!=\!0$, while the original image tokens and colour reference tokens are both set $t\!=\!1$. This positional encoding preserves spatial structure while facilitating semantic–chromatic alignment during attention computation.  

Unlike traditional colour-driven models that apply global tone transfer, {inspired by OmniControl~\cite{tan2025ominicontrol},}  ColourCrafter performs \textbf{token-level fusion} of image and colour features within the diffusion process.
The multi-modal attention mechanism then projects the fusion tokens into query Q, key K, and value V representations. It enables the computation of attention between all tokens:
\begin{equation}
\begin{split}
MMA([x;C_T;C_I])=softmax(\frac{QK^T}{\sqrt{d}})V,
\end{split}
\end{equation}
where $d$ is the embedding dimension. This selective fusion propagates colour only to semantically relevant regions, preserving texture and boundaries.  
LoRA-based fine-tuning~\cite{hu2022lora} is adopted for efficiency while freezing base parameters.  
The latent diffusion loss is:
\begin{equation}
\begin{split}
L_{diffusion}=&\mathbb{E}_{t,x,C_{I_o},C_{I_c},C_T}\![\|v_{\theta}(z_t,t,C_{I_o},C_{I_c},C_T)\\
&-(\epsilon-x)\|_2^2],
\end{split}
\end{equation}
where $z_t=(1-t)x+t\epsilon$, $\epsilon\!\sim\!\mathcal{N}(0,1)$.

\vspace{-10pt}
\subsubsection{Perceptual Lab-space Constraint}
\label{subsubsec:lab-space}

Although latent-space fusion preserves structural coherence, it operates in a feature domain misaligned with human perception, leading to residual colour bias. 
To address this, we impose a perceptual constraint in the CIE~Lab colour space~\cite{gomez2016comparison}, where distances more faithfully reflect human-perceived chromatic differences. 
Specifically, we introduce a \textbf{Lab-space Loss} that measures the discrepancy between the predicted image $I_{pred}$ and its ground truth $I_{gt}$ in this perceptually uniform space. 
Both images are decoded through the VAE and converted from RGB to Lab:
\begin{equation}
\begin{split}
& L_{pred},a_{pred},b_{pred}=\text{RGB2Lab}(I_{pred}),\\
& L_{gt},a_{gt},b_{gt}=\text{RGB2Lab}(I_{gt}),
\end{split}
\end{equation}
where \text{RGB2Lab} denotes the standard CIE transformation from RGB to the perceptually uniform Lab space. 
Here $L$ encodes luminance, while $a$ and $b$ capture opponent colour channels (green–red and blue–yellow), making Euclidean distance a perceptually meaningful measure of colour difference. 
The channel-wise mean-squared errors are then computed as:
\begin{equation}
\begin{split}
& L_L = \|L_{pred}-L_{gt}\|_2^2,\\
& L_{ab} = \|a_{pred}-a_{gt}\|_2^2 + \|b_{pred}-b_{gt}\|_2^2,
\end{split}
\end{equation}
combined as:
\begin{equation}
L_{Lab}=\lambda_L L_L + L_{ab}.
\end{equation}

A spatial mask $M\!\in\![0,1]^{H\times W}$ restricts this loss to edited regions.  
$M$ is derived from the \textbf{ColourfulSet} dataset by computing per-pixel colour differences between pre- and post-edited pairs (Sec.~\ref{subsec:dataset}).  
This focuses supervision on intended areas and avoids interference from unedited regions:
\begin{equation}
L_{Lab}^{mask}=M\odot L_{Lab},
\end{equation}
where $\odot$ denotes element-wise multiplication.  
The final objective is:
\begin{equation}
Loss=L_{diffusion}+ \lambda_{Lab} L_{Lab}^{mask},
\end{equation}
jointly enforcing latent-space structure and perceptual colour fidelity.

\vspace{-10pt}
\subsection{ColourfulSet Dataset}
\label{subsec:dataset}

To support model training and evaluation, we construct \textbf{ColourfulSet}, a large-scale dataset featuring continuous and locally controllable colour variations.  
From COCO2017~\cite{lin2014microsoft}, we select 1,200 images across 40 object categories suitable for colour editing.  
Using 144 CSS-standard colour names (\url{https://www.w3.org/TR/css-color-4/}), we generate prompts of the form “colour the [object] [colour]” and synthesise edited versions with the pre-trained Flux.1-Kontext model~\cite{lo2024distraction}.  


\begin{figure}[t]
    \centering
    \includegraphics[width=0.8\linewidth]{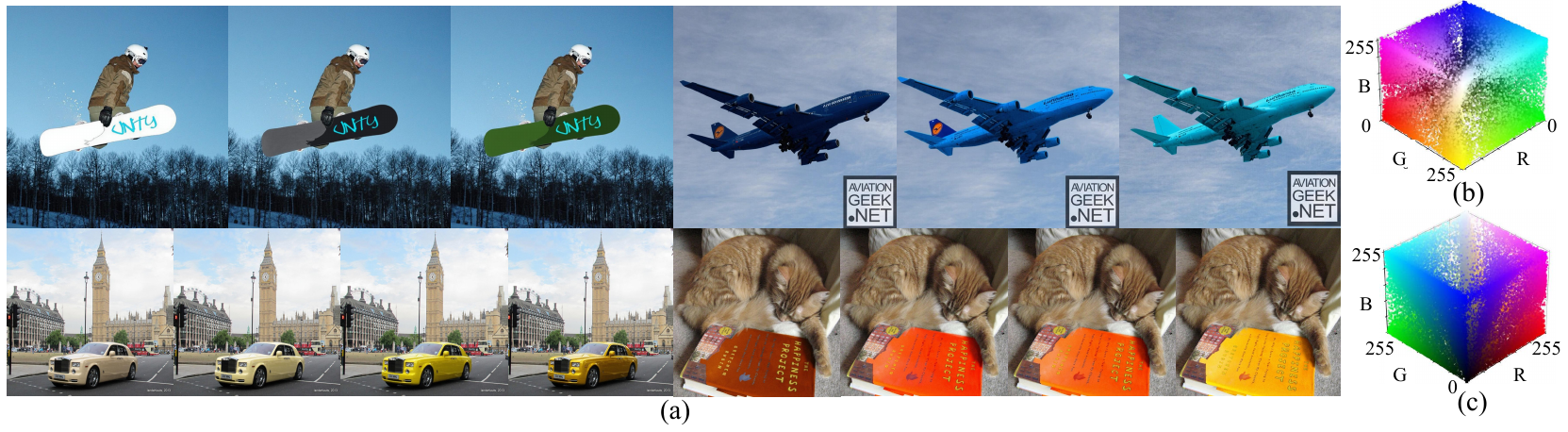}
    \vspace{-5pt}
    \caption{{\bf Examples from the ColourfulSet dataset.} 
    (a) shows edited images from different categories under various target colour references. 
    (b–c) Visualise the colour distribution of ColourfulSet from two perspectives in RGB space, demonstrating its diversity and uniform coverage.}
    \label{fig:dataset}
    \vspace{-20pt}
\end{figure}

Data quality is automatically verified by the Qwen model~\cite{qwen2.5-VL}, which scores each result by (i) task success, (ii) background consistency, and (iii) colour realism.  
Roughly 80,000 high-quality pairs are retained.  
For each pair, object masks are extracted with Segment-Anything Model (SAM)~\cite{ravi2024sam}, and pixel-wise colour difference is computed:
\begin{equation}
D(x,y)=\|I_e(x,y)-I_o(x,y)\|_2,
\end{equation}
where $I_o$ and $I_e$ denote the original and edited images.  
A percentile threshold $T_p$ isolates significant change regions; connected-component analysis removes noise, yielding the binary mask $M$ used for supervision.  
Within each mask, pixels are clustered in RGB space:
\begin{equation}
c_k=\frac{1}{N_k}\!\sum_{(x,y)\in C_k}\!I_e(x,y),
\end{equation}
where $C_k$ is the $k$-th cluster and $N_k$ its pixel count.  
The dominant cluster (largest $N_k$) defines the representative target colour $c_{target}=[R,G,B]$.  
This process ensures pixel-accurate colour–region alignment, providing a solid foundation for training and evaluating fine-grained colour editing.

These masks and colour annotations are also utilised during training to provide explicit region-level supervision for the Lab-space constraint, completing a closed loop between data generation and model optimisation.
{Fig.~\ref{fig:dataset} shows the examples from the ColourfulSet, please refer to our Supplementary Material for more examples.}

\vspace{-10pt}
\section{Experiments}
\label{sec:experiments}

\vspace{-5pt}
\subsection{Setting}
\label{subsec:setting}

\noindent\textbf{Implementation Details.}
We adopt Flux.1-Kontext~\cite{lo2024distraction} as the backbone and fine-tune it using LoRA~\cite{hou2025gencolor} with a rank of 16.  
Training uses the Adam optimiser with a learning rate of $3\times10^{-5}$, batch size 1, and a guidance scale of 3.5.  
All models are trained for 10,000 steps on a single NVIDIA A800 GPU, requiring about 40 hours.  
During inference, the denoising step is set to 28.  
Unless otherwise specified, the text prompt is \emph{``colour the [object]''}.


{\noindent\textbf{Comparison Methods.}
We compare ColourCrafter with:
(i) ColorPeel~\cite{butt2024colorpeel},
(ii) ColorBind~\cite{shomer2025color},
(iii) ControlColor~\cite{liang2025control},
(iv) Flux.1-Kontext~\cite{labs2025flux1kontextflowmatching},
(v) FlowEdit~\cite{kulikov2025flowedit}, and
(vi) UniEdit-Flow~\cite{jiao2025uniedit}.
ColorPeel, ColorBind and ControlColor adopt the same input settings as in the original paper. Flux.1-Kontext, FlowEdit and UniEdit-Flow takes the original image and a textual colour description, where we select the colour name closest to the target RGB reference. }

\begin{figure*}[t]
    \centering
    \includegraphics[width=0.8\linewidth]{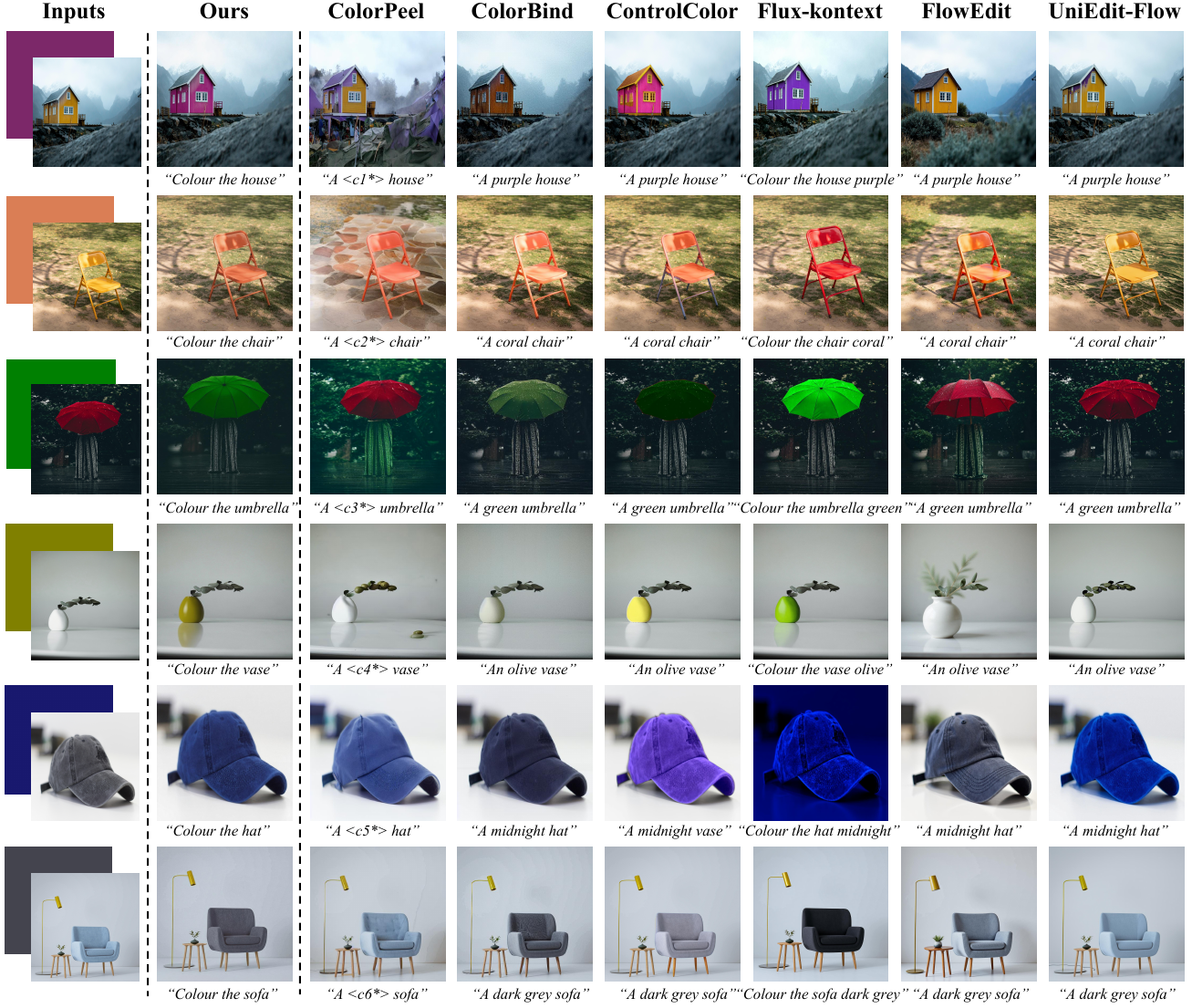}
    \vspace{-5pt}
        \caption{{\bf Comparison with other methods.} 
        The first column shows the reference colours and original images. 
        The comparison demonstrates that our method achieves more precise and fine-grained colour editing while preserving structural integrity and background consistency.}
    \label{fig:main result}
    \vspace{-25pt}
\end{figure*}

\noindent\textbf{Evaluation Data.}
Existing general editing datasets contain many objects unsuitable for colour editing.  
We therefore curate a dedicated benchmark from Unsplash (\url{https://unsplash.com/}),  
containing over 90 high-resolution ($1024\times1024$) images across 38 object categories.  
The dataset focuses on objects with well-defined structures and colours, facilitating fine-grained evaluation.

\noindent\textbf{Evaluation Metrics.}
For colour editing accuracy, we employ perceptual metrics in CIE Lab space:  
the Euclidean colour difference $\Delta E_{00}$~\cite{sharma2005ciede2000} and its chroma–hue variant $\Delta E_{Ch}$ (luminance removed).  
Additionally, we measure the mean angular error in sRGB space ($\mathrm{MAE}_{\mathrm{RGB}}$)  
and the hue angular error ($\mathrm{MAE}_{\mathrm{Hue}}$) to quantify chromatic deviation independent of luminance.  
For overall editing quality, we report Learned Perceptual Image Patch  Similarity (LPIPS)~\cite{zhang2018unreasonable} for perceptual similarity  between the edited and reference images,  
Structure Similarity Index Measure (SSIM)~\cite{wang2004image} for structural consistency,  
and Frechet Inception Distance (FID)~\cite{heusel2017gans} for realism and naturalness.

\vspace{-10pt}
\subsection{Results and Analysis}
\label{subsec:main result}

\subsubsection{Qualitative Comparison}
{Fig.~\ref{fig:main result} presents qualitative comparisons with existing colour editing methods. 
ColorPeel exhibits global colour rendering effects, often introducing saturation shifts and boundary artefacts due to the lack of explicit colour–region correspondence. 
Although ColorBind and ControlColor apply spatial masks to constrain editing regions, their colour conditioning remains globally injected in the latent space, leading to noticeable hue deviations within the masked areas and limited fine-grained chromatic precision. 
Flux.1-Kontext can alter object colours through textual guidance, yet its control is coarse and category-level, failing to reproduce subtle tone variations within the same colour family. 
FlowEdit and UniEdit-Flow show unstable colour mapping in several cases, revealing difficulties in maintaining consistent chromatic alignment across diverse scenarios.
In contrast, ColourCrafter establishes explicit colour–pixel correspondence through token-level fusion of RGB and image tokens. 
This enables selective, region-aware colour propagation while preserving material properties, illumination consistency, and background structure. 
The results demonstrate that bridging semantic localisation and chromatic alignment is essential for precise and structure-preserving recolouring.
We also compared with other General-purpose instruction editing model, such as qwen-image-edit~\cite{wu2025qwenimagetechnicalreport}, Flux.2~\cite{flux-2-2025}, please refer to the Supplementary Material for more comparison examples.}

\vspace{-15pt}
\begin{table}
\centering
\caption{{\bf Quantitative comparison on the colour editing task.}}
\vspace{-5pt}
\resizebox{0.73\linewidth}{!}{
\begin{tabular}{lccccccc}
\toprule
\multirow{2}{*}{\textbf{Method}} 
& \multicolumn{4}{c}{\textbf{Colour}} 
& \multicolumn{3}{c}{\textbf{Editing}} \\ 
\cmidrule(lr){2-5} \cmidrule(lr){6-8}
& $\Delta E_{00} \downarrow$  
& $\Delta E_{Ch} \downarrow$ 
& $\text{MAE}_{\text{RGB}} \downarrow$ 
& $\text{MAE}_{\text{Hue}} \downarrow$ 
& LPIPS $\downarrow$ 
& SSIM $\uparrow$ 
& FID $\downarrow$ \\ 
\midrule
ColorPeel~\cite{butt2024colorpeel}                 & 60.96 & 53.40 & 77.79 & 61.95 & 0.38 & 0.72 & 79.30 \\
ColorBind~\cite{shum2025color}                 & 53.42 & 47.26 & 62.78 & 44.12 & 0.36 & 0.70 & 69.68 \\
Control-Color~\cite{liang2025control}             & 32.82 & 28.44 & 71.90 & 20.31 & 0.28 & 0.82 & 63.87 \\
Flux.1-Kontext~\cite{labs2025flux1kontextflowmatching}            & 41.36 & 36.31 & 48.34 & 23.19 & 0.29 & 0.74 & 63.01 \\
FlowEdit~\cite{kulikov2025flowedit}                  & 66.26 & 56.32 & 83.85 & 61.59 & 0.29 & 0.79 & 81.32 \\
UniEdit-Flow~\cite{jiao2025uniedit}              & 73.26 & 63.87 & 91.60 & 79.47 & 0.31 & 0.80 & \textbf{46.57} \\
\midrule
\rowcolor{gray!15}
\textbf{Ours}             & \textbf{28.52} & \textbf{23.55} & \textbf{35.84} & \textbf{14.88} & \textbf{0.28} & \textbf{0.83} & 51.91 \\
\bottomrule
\end{tabular}
}
\label{tab: main table}
\vspace{-30pt}
\end{table}


\vspace{-10pt}
\subsubsection{Quantitative Comparison}
The quantitative results in Tab.~\ref{tab: main table} further confirm the superiority of ColourCrafter.  
Our method consistently achieves the lowest colour errors ($\Delta E_{00}$, $\Delta E_{Ch}$, $\mathrm{MAE}_{\mathrm{RGB}}$, $\mathrm{MAE}_{\mathrm{Hue}}$)  
and competitive or better perceptual scores (LPIPS $\downarrow$, SSIM $\uparrow$, FID $\downarrow$) compared with all methods.  
This demonstrates that ColourCrafter effectively balances chromatic precision with structural and perceptual quality.

\begin{figure*}
    \centering
    \includegraphics[width=0.85\linewidth]{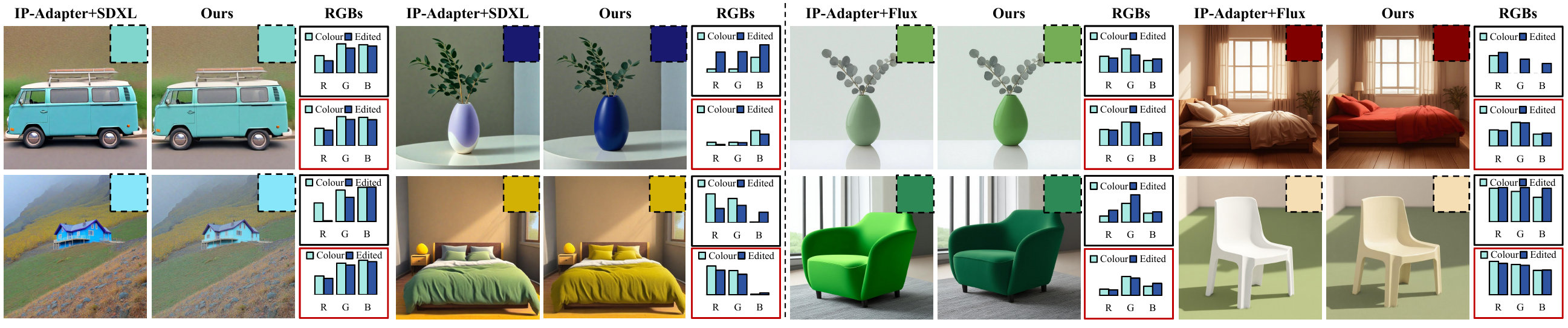}
    \vspace{-5pt}
    \caption{{\bf Comparison with IP-Adapter methods.} In the RGB column,
black and red boxes indicate the target object colour and the reference colour for results produced by the IP-Adapter and ColourCrafter (Ours), respectively.}
    \label{fig:ip-adapter}
    \vspace{-20pt}
\end{figure*}

\vspace{-15pt}
{\subsubsection{Comparison with IP-Adapter Methods}
\label{subsubsec:ip-adapter}
IP-Adapter is widely adopted for image-based conditioning. 
To evaluate ColourCrafter under local fine-grained colour control, we compare it with (i) IP-Adapter + SDXL~\cite{rowles2024ipadapter} and (ii) IP-Adapter + Flux~\cite{flux2024}. 
As shown in Fig.~\ref{fig:ip-adapter}, IP-Adapter injects global image-level features into the diffusion process. 
While this mechanism transfers overall visual style, it does not establish explicit colour–pixel correspondence for the target object. 
Consequently, the edited results often exhibit hue deviation from the reference colour, even within the intended region.
The RGB histograms further reveal this discrepancy. 
For IP-Adapter, the object colour distribution (black boxes) significantly deviates from the reference RGB values, indicating unstable chromatic alignment. 
In contrast, ColourCrafter achieves close alignment between the target object colour and the reference colour (red boxes), demonstrating precise colour–region correspondence.
These results confirm that global feature injection is insufficient for fine-grained recolouring, and explicit token-level chromatic alignment is essential for accurate local colour control.}

\vspace{-15pt}

\vspace{-5pt}
\subsubsection{Smooth and Progressive Colour Change}
Meanwhile, as shown in Fig.~\ref{fig:fig1}, our method exhibits no abrupt colour transitions during the editing process. When the reference RGB colours vary in an arithmetic progression, the generated results display continuous and uniform colour transitions within the edited regions, demonstrating strong colour smoothness and consistency. This indicates that our model maintains stable responses in the colour space and enables progressive control over colour variations.

\vspace{-15pt}
\begin{figure}
\begin{minipage}[t]{0.55\linewidth}
    \centering
    \includegraphics[width=0.85\linewidth]{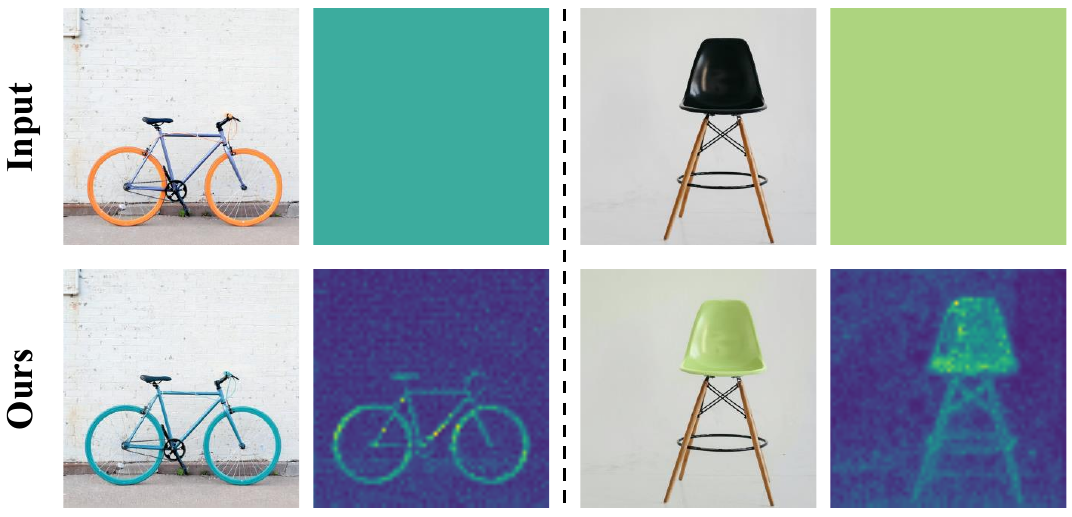}
    \captionof{figure}{{\bf Attention maps conditioned on RGB reference images.} 
    The maps highlight semantically relevant regions that correspond to the target colours, demonstrating the model’s ability to achieve accurate and localised colour editing.}
    \label{fig:attention}
\end{minipage}
\begin{minipage}[t]{0.43\linewidth}
    \centering
    \includegraphics[width=0.9\linewidth]{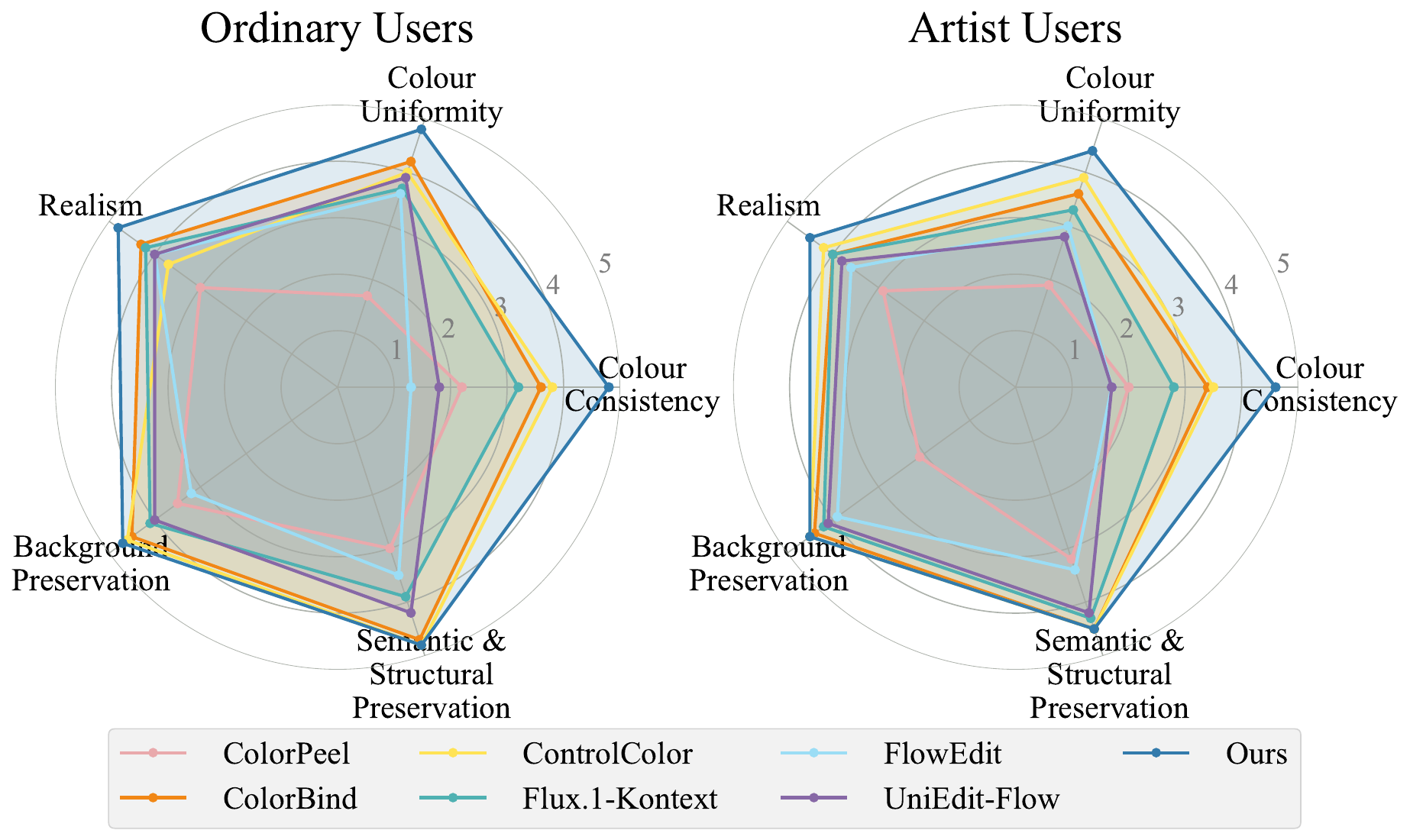}
    \captionof{figure}{{\bf Results of the user study.} 
    The left panel shows ratings from ordinary users, and the right panel shows ratings from professional artists.}
    \label{fig:userstudy}
\end{minipage}
\vspace{-25pt}
\end{figure}

\vspace{-15pt}
\subsubsection{Attention Visualization Analysis}
\label{subsubsec: attention}
As shown in Fig.~\ref{fig:attention}, we visualize the attention maps conditioned on the colour images to investigate how the model utilizes colour information during the editing process. The results reveal that our model automatically focuses on image regions that are semantically correlated with the target colour. For instance, in the bicycle example, the attention is concentrated on the wheel area, which precisely aligns with the cyan reference colour, while in the chair example, the model accurately attends to the seat region and successfully performs the black-to-green colour transformation. These observations indicate that our method effectively captures spatially localized correspondences between the input colour image and the target region, ensuring accurate and consistent colour transfer.

\vspace{-10pt}
\subsubsection{User Study}
\label{subsubsec: user study}
We further conduct a user study involving two groups: general users and professional artists.  
Participants rate colour Consistency, colour Uniformity, Semantic \& Structural Preservation, Background Preservation, and overall visual Realism on a 1–5 Likert scale. 
As shown in Fig.~\ref{fig:userstudy}, ColourCrafter is consistently preferred across all metrics,  
confirming its superior perceptual quality and naturalness.

\begin{figure}
\begin{minipage}[t]{0.63\linewidth}
    \centering
    \includegraphics[width=0.92\linewidth]{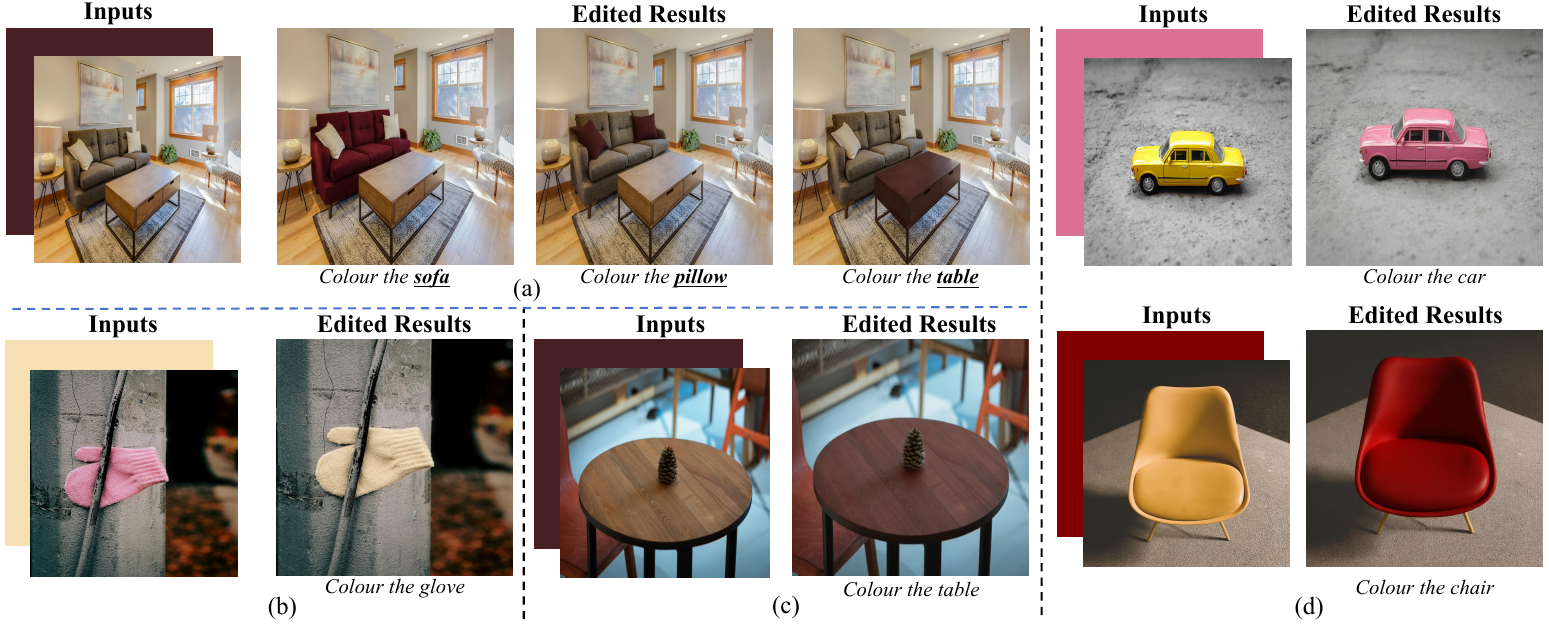}
    \vspace{-5pt}
    \captionof{figure}{{\bf Complex scenarios results.} (a) Multiple objects, (b) Occlusion, (c) Complex textures, (d) Lighting conditions.}
    \label{fig:complex}
\end{minipage}
\begin{minipage}[t]{0.32\linewidth}
    \centering
    \includegraphics[width=0.65\linewidth]{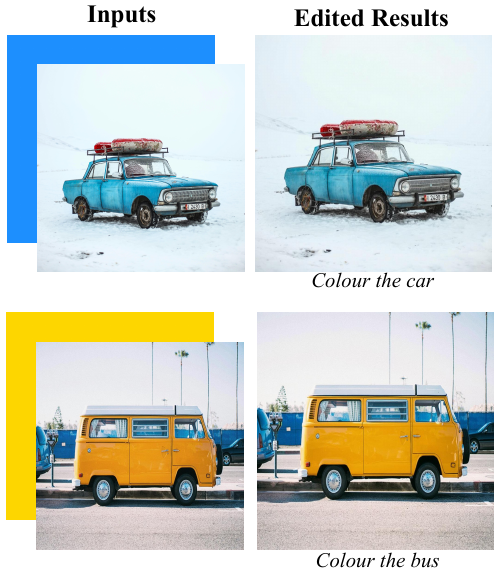}
    \vspace{-5pt}
    \captionof{figure}{{\bf Results of failure cases.} }
    \label{fig:failure_case}
\end{minipage}
\vspace{-20pt}
\end{figure}

\vspace{-10pt}
{
\subsubsection{Performance in Complex Scenarios} \label{subsubsec:complex} We further evaluate ColourCrafter under challenging real-world scenarios, including multiple objects, occlusions, complex textures, and diverse lighting conditions (Fig.~\ref{fig:complex}). In scenes containing multiple objects, ColourCrafter correctly identifies and recolours only the specified target, avoiding unintended colour propagation to semantically similar regions. Under occlusion, the model maintains consistent chromatic alignment on partially visible objects, indicating stable colour–pixel correspondence. For objects with intricate textures or material patterns, ColourCrafter preserves structural details while accurately adjusting chrominance, demonstrating structure-aware colour propagation. Moreover, under varying illumination conditions, the edited results retain realistic shading and brightness, reflecting the effectiveness of luminance–chrominance decoupling in the Lab-space loss.}

\vspace{-10pt}
{
\subsubsection{Failure Cases}
\label{subsubsec:failure_case}
As shown in Fig.~\ref{fig:failure_case}, we observe that when the target colour is close to the object’s original colour (e.g., $\Delta E_{00} < 35.7$, from statistical analysis of failure cases), the editing results are more prone to colour deviations. This is because when the colour difference is small, the model receives a weak colour-change signal, making it difficult to form sufficiently strong gradient guidance during the diffusion denoising process, which consequently leads to insufficient editing strength or a shift towards neighbouring colour distributions.
}

\vspace{-18pt}

\vspace{-15pt}
\subsection{Ablation study}
\label{subsec: ablation}

\noindent\textbf{Effect of Luminance Weight in Lab Loss}
To investigate the effect of the luminance component in colour editing, we perform an ablation study on the weight $\lambda_L$ in the Lab colour space loss, as shown in Tab.~\ref{tab:ablation_L}. When $\lambda_L = 0$ , the model relies solely on the a and b channels, leading to reasonable hue accuracy ($\text{MAE}_{\text{RGB}}$) but degraded overall colour consistency ($\Delta E_{00}$ and $\Delta E_{ab}$). As $\lambda_L$ increases, the colour reconstruction quality improves, indicating that an appropriate luminance constraint enhances the stability and perceptual coherence of colour editing. However, excessively large $\lambda_L$ values (e.g., $\lambda_L = 0.8$) overly emphasize brightness, causing a loss of colour saturation. Therefore, we set $\lambda_L = 0.5$ as the optimal balance between luminance and chromatic contributions in our experiments.

\begin{figure*}[t]
    \centering
    \includegraphics[width=0.85\linewidth]{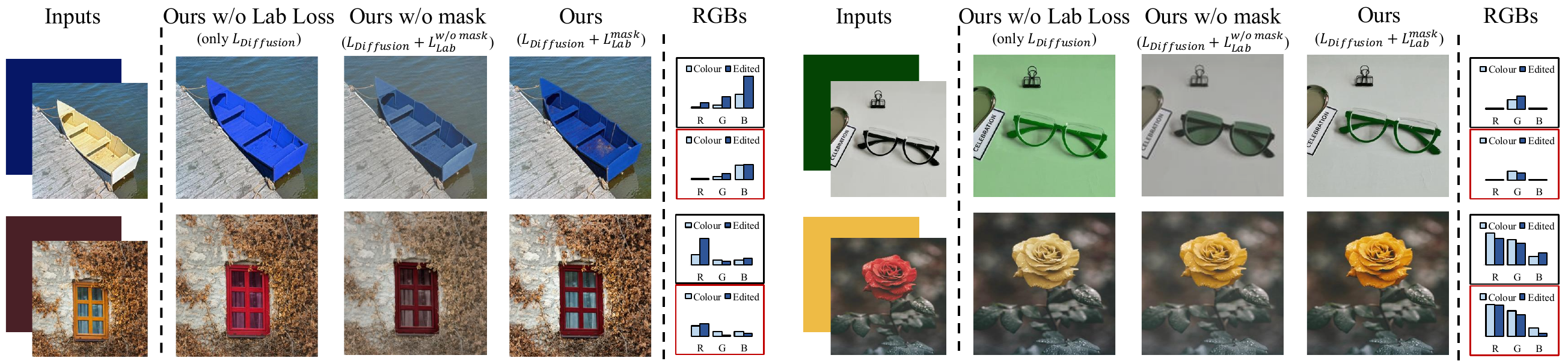}
    \vspace{-5pt}
    \caption{{\bf Qualitative ablation of different loss functions.} 
    In the RGB column, black and red boxes indicate the target object colour and the reference colour 
    for results produced by the Ours w/o Lab Loss and ColourCrafter (Ours), respectively.}
    \label{fig:baseline}
    \vspace{-15pt}
\end{figure*}

\begin{table}
\centering
\begin{minipage}[t]{0.48\linewidth}
\centering
\caption{\textbf{Ablation study over $\lambda_L$}. Best in bold.}
\vspace{-5pt}
\resizebox{0.85\linewidth}{!}{%
\begin{tabularx}{1.2\linewidth}{>{\centering\arraybackslash}p{0.4cm} *{4}{>{\centering\arraybackslash}X}}
\toprule
{\textbf{$\lambda_L$}} & $\Delta E_{00}\downarrow$ & $\Delta E_{Ch}\downarrow$ & $\text{MAE}_{\text{RGB}}\downarrow$ & $\text{MAE}_{\text{Hue}}\downarrow$ \\ 
\midrule
0    & 45.15  & 39.18  & 56.06  & 34.47 \\
0.2  & 35.34  & 29.96  & 44.39  & 21.36 \\
\rowcolor{gray!15}
0.5  & \textbf{34.54}  & 29.51  & \textbf{42.70} & \textbf{18.49} \\
0.8  & 40.79  & 33.93  & 51.37  & 23.37 \\
1    & 35.19  & \textbf{27.87}  & 50.09  & 23.10 \\
\bottomrule
\end{tabularx}
}
\label{tab:ablation_L}
\end{minipage}
\hfill
\begin{minipage}[t]{0.48\linewidth}
\centering
\caption{\textbf{Ablation study over $\lambda_{Lab}$}. Best in bold.}
\vspace{-5pt}
\resizebox{0.85\linewidth}{!}{%
\begin{tabularx}{1.2\linewidth}{>{\centering\arraybackslash}p{0.4cm} *{4}{>{\centering\arraybackslash}X}}
\toprule
{\textbf{$\lambda_{Lab}$}} & $\Delta E_{00} \downarrow$ & $\Delta E_{Ch} \downarrow$ & $\text{MAE}_{\text{RGB}} \downarrow$ & $\text{MAE}_{\text{Hue}} \downarrow$ \\ 
\midrule
0     & 32.87 & 26.09 & 48.59 & 20.48 \\
0.3   & 30.36 & 24.42 & 43.20 & 20.02 \\   
0.5   & 34.54 & 29.51 & 42.70 & 18.49 \\
\rowcolor{gray!15}
0.8   & \textbf{28.53} & \textbf{23.55} & \textbf{35.84} & \textbf{14.88} \\
1     & 37.50 & 31.67 & 47.34 & 20.72 \\      
\bottomrule
\end{tabularx}
}
\label{tab:ablation_Lab}
\end{minipage}
\vspace{-20pt}
\end{table}

\noindent\textbf{Effect of the Lab Loss Weight}
\label{subsubsec:Lab}
To evaluate the contribution of the Lab Loss in the overall optimization, we conduct an ablation study on the weighting factor $\lambda_{Lab}$ that balances the Lab Loss and Diffusion Loss, as shown in Tab.~\ref{tab:ablation_Lab}. When $\lambda_{Lab}=0$, the model is trained solely with Diffusion Loss, resulting in noticeably higher $\Delta E_{00}$ and $\Delta E_{ab}$, which indicates inaccurate colour reproduction due to the absence of explicit colour supervision. As $\lambda_{Lab}$ increases, the colour accuracy steadily improves, with lower $\text{MAE}_{\text{RGB}}$ and $\text{MAE}_{\text{Hue}}$ values, suggesting that the Lab Loss effectively enhances the precision and stability of colour transfer. However, excessively large $\lambda_{Lab}$ values overly constrain the generative process, reducing colour richness and texture fidelity. Therefore, we set $\lambda_{Lab}=0.8$ in our final model to achieve an optimal balance between colour accuracy and perceptual realism.

Meanwhile, as shown in Fig.~\ref{fig:baseline}, we compare the generation results between the {Ours w/o Lab Loss} and our method. The Ours w/o Lab Loss relies solely on the generative capability of the diffusion model for colour adjustment, lacking explicit colour constraints; therefore, it can only achieve coarse-grained colour control, often leading to deviations in brightness and chromaticity. After introducing the Lab Loss, the model acquires a more precise ability for colour transfer. Compared with the Baseline, the model with Lab Loss shows significant improvements in both colour consistency and alignment with the target colour. The RGB histograms further validate the effectiveness of the Lab Loss: the colour distribution of our method is closer to the target RGB values, demonstrating higher colour mapping accuracy and perceptual consistency.

\noindent\textbf{Effect of Mask Weighting}
Recall that $M$ originates from ColourfulSet’s colour-change annotations.
However, as illustrated in Fig.~\ref{fig:baseline}, the Ours w/o mask setting applies the Lab Loss over the entire image without mask weighting. This global constraint causes the model to balance the colour difference between edited and unedited regions, leading to a decrease in overall saturation and a slightly faded or whitish appearance. In contrast, our full model (Ours) introduces mask weighting in the Lab Loss, restricting colour supervision to the intended editing region. This spatially selective constraint enables precise local colour control while preserving background consistency.



\begin{figure}
    \centering
    \includegraphics[width=0.75\linewidth]{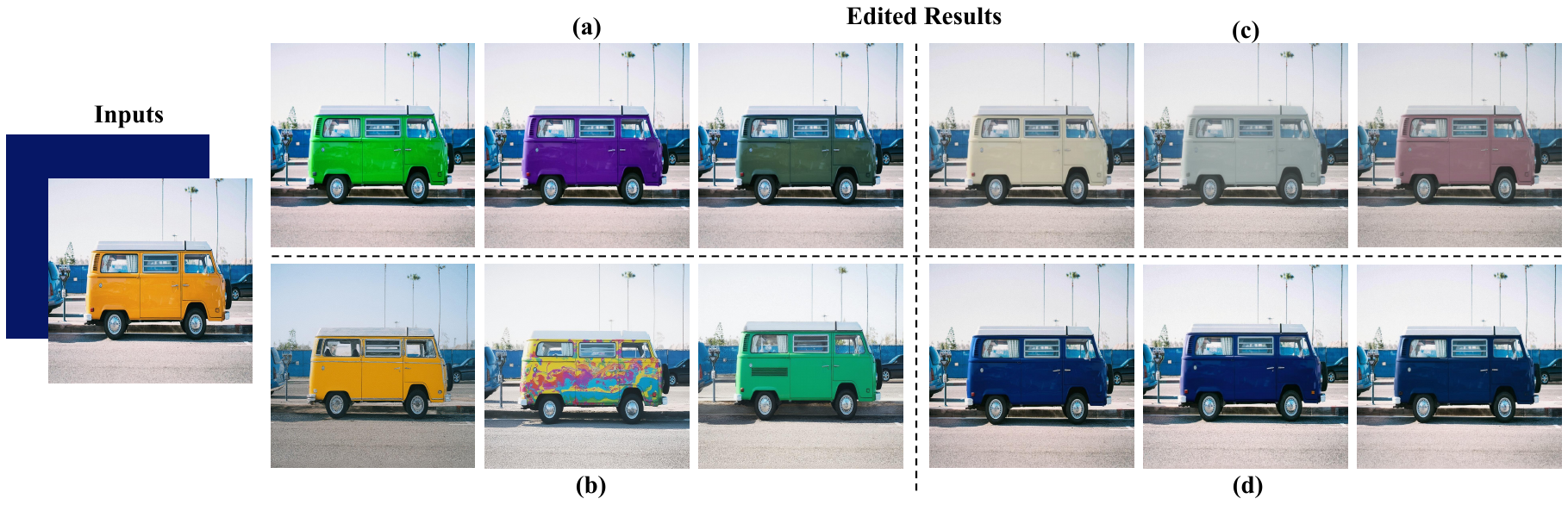}
    \vspace{-5pt}
    \caption{{\bf Results of different injection methods and colour constraints methods under different random seeds.} (a) denotes encoding RGB colour patches as text tokens, (b) denotes encoding RGB arrays as text tokens, (c) denotes RGB Loss constraint, (d) denotes Ours.}
    \label{fig:injection}
    \vspace{-20pt}
\end{figure}

\noindent\textbf{Why not inject colour information into text token?}
We explore different injection strategies: (a) encoding RGB colour patches as text tokens; (b) encoding RGB arrays as text tokens. As shown in Fig.~\ref{fig:injection}, the generated results exhibit highly random colour variations under different random seeds, indicating that the model fails to establish a stable mapping between the input colour conditions and the generated outputs. When colour information is injected via text tokens, it is inherently misaligned with the semantic space, causing the colour condition signal to be progressively weakened during the diffusion denoising process.

\noindent\textbf{Why not use RGB Loss constraint?}
We compare ColourCrafter with RGB loss, as shown in Fig.~\ref{fig:injection} (c), the RGB loss acting as a global statistical constraint, tends to drive the model towards the mean colour distribution, resulting in desaturated, greyish outputs and a lack of spatial discrimination.

\vspace{-15pt}

\vspace{-15pt}
\section{Limitations}
\label{sec:limitations}
\vspace{-10pt}

Although ColourCrafter demonstrates precise and structure-preserving colour editing, several aspects warrant further improvement.  
(1) The model is trained on synthetic colour variations generated by Flux.1-Kontext; such data, while large-scale, may not capture the full complexity of human colour perception or artistic style.  
(2) The current training pipeline depends on paired supervision between colour references and target images, which limits scalability to real-world or unpaired editing scenarios.  
{(3) Our method currently supports editing only a single object with a single colour at a time; editing multiple objects or multiple colours requires multiple sequential editing steps.}

\vspace{-10pt}
\section{Conclusion}
\label{sec:conclusion}
\vspace{-10pt}

We introduced \textbf{ColourCrafter}, a unified diffusion framework for fine-grained and structure-preserving colour editing.  
By performing token-level fusion of semantic and chromatic cues, and by introducing a perceptual Lab-space constraint, ColourCrafter achieves accurate, controllable, and perceptually stable recolouring across diverse categories.  
The proposed \textbf{ColourfulSet} dataset provides large-scale, continuous, and locally controllable colour variations, supporting both model training and fair evaluation.  
Our results highlight the effectiveness of integrating semantic localisation and continuous chromatic precision, and we hope this study inspires future research on controllable and perceptually aligned image generation.



%
%


\bibliographystyle{splncs04}
\bibliography{main}

\clearpage
\end{document}